\documentclass[preprint,12pt]{elsarticle}

\usepackage{amssymb}
\usepackage{hyperref}
\usepackage{subcaption}
\usepackage{graphicx}
\usepackage{caption}
\usepackage[table]{xcolor}
\usepackage{booktabs}
\usepackage{siunitx}   
\usepackage{tabularx}
\usepackage{makecell}
\usepackage{arydshln} 
\usepackage{xcolor}
\usepackage{soul}

\usepackage{amsmath}
\usepackage{multirow}
\usepackage{multicol}
\usepackage{adjustbox}
\usepackage{lscape}

\journal{Applied Soft Computing}

\begin{document}

\begin{frontmatter}

\title{Effect of Patch Size on Fine-Tuning Vision Transformers in Two-Dimensional and Three-Dimensional Medical Image Classification}

\author[label1]{Massoud Dehghan}\ead{massoud.dehghan@dp-uni.ac.at}
\author[label1]{Ramona Woitek}\ead{ramona.woitek@dp-uni.ac.at}
\author[label1]{Amirreza Mahbod\fnref{equal3}}\ead{amirreza.mahbod@dp-uni.ac.at}

\fntext[equal3]{Corresponding Author}

\affiliation[label1]{organization={Research Center for Medical Image Analysis and Artificial Intelligence, Department of Medicine, Faculty of Medicine and Dentistry, Danube Private University},
            city={Krems an der Donau},
            postcode={3500},
            country={Austria}}




\begin{abstract}
Vision Transformers (ViTs) and their variants have become state-of-the-art in many computer vision tasks. While substantial research has focused on architectural improvements, the impact of patch size, a crucial design choice in ViTs, remains underexplored, particularly in medical domains where both two-dimensional (2D) and three-dimensional (3D) imaging modalities exist.

In this study, using 12 medical imaging datasets from various imaging modalities (including seven 2D and five 3D datasets), we conduct a thorough evaluation of how different patch sizes affect ViT classification performance. Using a single graphical processing unit (GPU) and a range of patch sizes (1, 2, 4, 7, 14, 28), we fine-tune ViT models and observe consistent improvements in classification performance with smaller patch sizes (1, 2, and 4), which achieve the best results across nearly all datasets. More specifically, our results indicate improvements in balanced accuracy of up to 12.78 percentage points for 2D datasets (patch size 2 vs. 28) and up to 23.78 percentage points for 3D datasets (patch size 1 vs. 14), at the cost of increased computational expense. Moreover, by combining the models trained with patch sizes 1, 2, and 4 using four fusion strategies, we achieve further performance gains in most cases, particularly for the 2D datasets. We further investigate the generalizability of these findings across higher image resolutions, different transformer architectures, and varying amounts of training data, observing performance trends similar to those of the main experiments. Our full implementation, along with guidelines for running the code, is publicly available on GitHub:~\url{https://github.com/HealMaDe/MedViT}.



\end{abstract}




\end{frontmatter}



\section{Introduction}
\label{sec:introduction}

The broad range of medical imaging modalities, such as magnetic resonance imaging (MRI), computed tomography (CT), dermatoscopy, X-ray, and ultrasound, which form the backbone of modern medicine, has created a noticeable demand for automated analysis systems for downstream tasks such as detection, segmentation, and classification~\cite{10.3389/fpubh.2023.1273253, handa2024wcebleedgen}. These systems can help reduce diagnostic errors made by medical experts, as they can process large volumes of data efficiently and learn patterns present in the data~\cite{ref1, ref2}. Among the various tasks defined in medical computer vision, image classification is one of the fundamental tasks, where the aim is to assign a label to an entire image, for example, benign versus malignant~\cite{7312934}.

While many automatic methods have been proposed for medical image classification, deep learning techniques, particularly convolutional neural networks (CNNs), have shown highly promising results over the past decade compared to traditional image processing or classical machine learning (ML) approaches. Popular CNN-based models such as ResNet~\cite{He2016}, DenseNet~\cite{huang2017densely}, EfficientNet (V1 and V2)~\cite{tan2019efficientnet, Tan2021}  and ensemble CNNs~\cite{mahbod2021pollen} have set new benchmarks in a wide range of medical and non-medical image classification tasks~\cite{He2016,ref3,huang2017densely,tan2019efficientnet}. These models learn hierarchical image features automatically without the need for manual feature extraction, which is one of their key advantages over classical ML methods that rely on potentially suboptimal hand-crafted features~\cite {annurev:/content/journals/10.1146/annurev-bioeng-071516-044442}. 

Convolutional kernels are effective feature extractors within their local receptive fields, but the inherent locality bias of CNNs limits their capacity to capture long-range dependencies in images. As a newer alternative to traditional convolutional architectures, vision transformers (ViTs)~\cite{ref8} rely on self-attention mechanisms that directly model global relationships between image patches, enabling a more flexible and expressive representation of contextual information~\cite{10.5555/3295222.3295349, AZAD2024103000}. ViTs were originally inspired by transformer architectures~\cite{10.5555/3295222.3295349} developed for natural language processing, and were adapted to vision tasks such as image classification through techniques including patch embedding and positional encoding~\cite{ref8}. Several studies have shown that ViT-based architectures outperform CNNs in various medical imaging tasks, demonstrating improved generalization and robustness across modalities~\cite{ref10, ref11, Takahashi2024}. Due to their excellent performance, many large-scale medical and non-medical foundation models, such as MedCLIP~\cite{wang2022medclip}, BioMedCLIP~\cite{zhang2023large}, UNI~\cite{Chen2024} and OpenCLIP~\cite{ilharco_gabriel_2021_5143773}, use ViTs as their backbone architecture, and these models have been widely adopted for downstream clinical and computational applications~\cite{MAHBOD2025110571, saeidi2024leveraging, bioengineering12121332}.

As stated in~\cite{ref8}, in a ViT, an input image is first split into fixed-size patches. Each patch is then flattened and transformed into patch embeddings. A learnable class token is added to the start of the sequence, and positional embeddings are included to keep spatial information. The sequence goes through several transformer encoder layers, which include multi-head self-attention and feed-forward networks. This setup enables interaction of global features among patches. Finally, the output of the final ViT encoder block is connected to a classification head (e.g., a multi-layer perceptron) to predict the image label.

Since the introduction of the first ViT models, several novel and modified approaches, such as Swin Transformers (V1 and V2)~\cite{ref12, 9879380}, and hybrid CNN-Transformer architectures~\cite{ref13,ref14, torbati2025acs} have been proposed to boost performance in visual recognition tasks. However, one fundamental component of ViT pipelines that is rarely discussed in the literature is the patching strategy used to tokenize the input images. Most ViT-based methods rely on a fixed patch size (commonly 14$\times$14 or 16$\times$16), even though patch size directly determines the number of tokens and, therefore, the model’s ability to capture fine-grained spatial information. In this study, our goal is to specifically examine the impact of different patch sizes on the overall performance of ViT for medical image classification. To the best of our knowledge, very few studies have explored ViTs from this perspective, particularly in the context of medical imaging, where both 2D and 3D datasets differ structurally from natural images.

Than et al. performed a preliminary evaluation of ViTs for classifying COVID-19 and lung diseases. They used various patch sizes, ranging from 16$\times$16 to 256$\times$256. Their findings indicated that choosing the right patch size can greatly affect classification performance, with larger patches leading to a noticeable drop in accuracy. However, they did not explore what happened to the performance if the size of the patches decreased~\cite{ref15}. In another study carried out by Liu et al., eliminating unnecessary input patches was examined to reduce computational cost while maintaining performance on high-resolution medical image classification tasks, although the systematic effect of patch sizes alone was not tested \cite{10030183}. Similarly, without considering the effect of various patch sizes, Halder et al. showed that ViT models can achieve state-of-the-art performance across diverse 2D biomedical imaging tasks using the MedMNIST V2 collection~\cite {ref17, Yang2023}. Additionally, Beyer et al. proposed FlexiViT, which is based on randomized patch sizes and adaptive positional embedding, and achieved better results compared to fixed patch sizes. However, the impact of specific patch-size choices was not discussed in their work~\cite{10205121}. In another study, Mantes et al. proposed MuViT, which jointly encodes views of the same image at multiple resolutions within a single ViT encoder by embedding all patches in a shared world-coordinate system with rotary positional embeddings~\cite{Mantes_2026_CVPR}. However, their work focuses on multi-scale microscopy analysis, primarily semantic segmentation, and uses a fixed patch size throughout, without examining the effect of patch size, which is the focus of our study. Finally, two research efforts closely related to our work investigated the effect of reducing ViT patch sizes on performance. Nguyen et al. showed that in a supervised learning setting, ViTs can benefit from smaller patch sizes while keeping the input image size fixed~\cite{ref18}. Similarly, Wang et al. reported consistent improvements in performance as patch sizes were reduced, revealing a scaling law in patchification in which accuracy continues to increase as patch size decreases~\cite{ref19}. However, both studies focused solely on 2D natural images and did not consider either 2D or 3D medical images. Moreover, the models in these studies were trained from scratch, which is often not feasible for small-scale medical imaging datasets, where fine-tuning is generally the preferred approach~\cite{Bahadir2024}.

In summary, there remains a clear gap in the literature regarding the systematic investigation of patch-size effects on medical image data. Most existing ViT-based studies have been developed and evaluated using natural image datasets rather than medical images. In addition, these models are often trained from scratch instead of being fine-tuned on domain-specific data. Current research also focuses almost entirely on 2D images, even though many medical imaging modalities, such as MRI and CT, are inherently 3D and require volumetric modeling for accurate diagnosis. Moreover, training of large-scale ViT models in previous studies has frequently relied on multiple high-end graphical processing units (GPU) or tensor processing units (TPU) clusters and substantial computational resources (for example, 50,000 A100 GPU hours in~\cite{ref19} or 10,000 TPU-core days in~\cite{ref20}), creating a significant computational barrier for many research groups.

In contrast, our study tackles all these limitations. We provide the first complete evaluation of how progressively smaller patch sizes affect ViT-based classification performance in both 2D and 3D medical imaging modalities. Importantly, we show that detailed patch-size analysis is possible on a single, modest GPU, as long as the appropriate dataset is used. Moreover, we show that fusing models trained with small patch sizes can further improve classification performance, and we examine the generalizability of our findings across higher image resolutions, different transformer architectures, and varying amounts of training data. These findings emphasize the practicality of conducting detailed tokenization studies in real-world research settings and offer new insights into the best patch-size selection for medical image analysis.

The main contributions of this study can be summarized as follows:

\begin{itemize}

\item We perform a thorough evaluation of the impact of patch size on the classification performance of ViTs.

\item We focus on fine-tuning ViTs, a common approach in the medical domain, and include both 2D and 3D medical imaging datasets using a single GPU setup in our experiments.

\item We show that fusing models trained with small patch sizes using different fusion strategies can further improve classification performance.

\item We further examine the generalizability of our findings across higher image resolutions, different transformer architectures, and varying amounts of training data.

\item Our implementation is publicly available, which supports transparency and reproducibility of our results.

\end{itemize}

In the rest of the paper, we explain our methodology and implementation setup in Section~\ref{sec:methodology}, report and discuss the results in Section~\ref{sec:results}, and conclude the work in Section~\ref{sec:conclusion}.

\section{Methodology}
\label{sec:methodology}

In ViTs, the computational complexity of the multi-head self-attention mechanism grows quadratically with the number of input tokens ($T$). So, the self-attention cost per layer is proportional to $O(T^2 d)$, where $d$ is the embedding dimension \cite{ref21}. 

To analyze how patch size affects computational cost ($C$), we compare a standard patch size of $p\times\!p$ with a smaller patch size of $\frac{p}{N}\times\frac{p}{N}$. For a 2D image of size $H\times\!W$, reducing the patch size raises the number of patches (tokens) from $T_1 = \frac{HW}{p^{2}}$ to $T_2 = N^{2}T_1$, which shows a growth factor of $N^2$. Since the self-attention operation scales with the square of the number of tokens, the attention cost per layer goes up from $C_1\propto T_1^{2}$ to $C_2\propto (N^{2}T_1)^{2} = N^{4}T_1^{2}$. As a result, the self-attention computation cost increases by a factor of  $N^{4}$. Extending this reasoning to 3D inputs with volume $H\times\!W\times\!D$ and 3D patches of size $p\times\!p\times\!p$, reducing the patch size to $\frac{p}{N}$ in each dimension increases the number of tokens from $T_1 = \frac{HWD}{p^{3}}$ to $T_2 = N^{3}T_1$, a growth factor of $N^{3}$. Consequently, the self-attention cost rises from $C_1 \propto T_1^{2}$ to $C_2 \propto (N^{3}T_1)^{2} = N^{6}T_1^{2}$. Therefore, in 3D ViTs, decreasing the patch size by a factor of $N$ in each dimension increases the self-attention computation cost by $N^{6}$, demonstrating the severe scaling challenges introduced by finer 3D tokenization. For example, halving the patch size (from $p$ to $\frac{p}{2}$) multiplies the self-attention cost by $16\times$ in 2D, and by $64\times$ in 3D due to cubic token growth. This steep increase in computational cost explains why smaller patches, although potentially beneficial for fine-grained feature extraction (as shown in the results section), have not been thoroughly investigated in the literature. 

\begin{figure}[t]
    \centering
    \includegraphics[width=\textwidth]{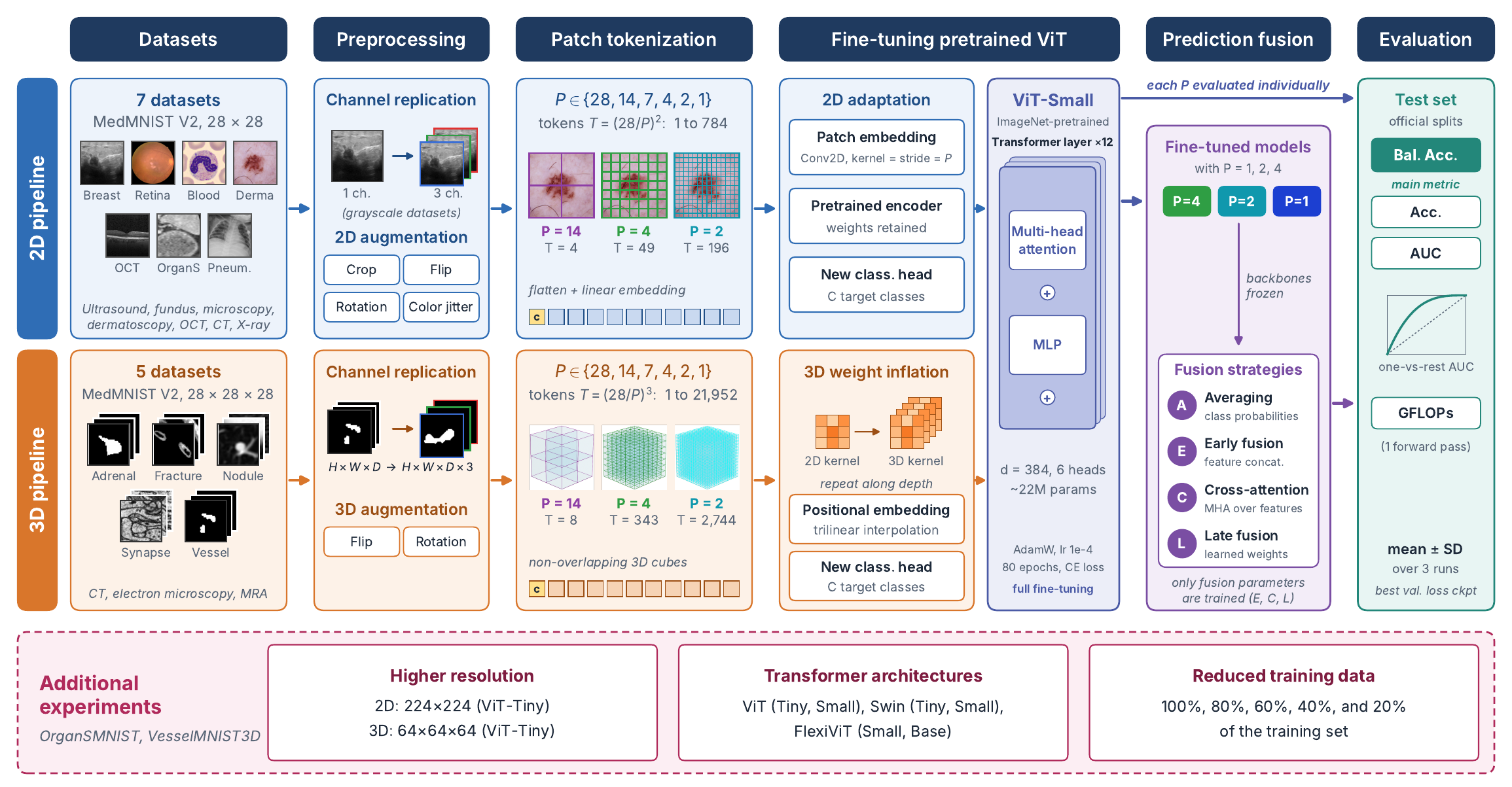}
   \caption{Overview of the proposed pipeline. 2D and 3D MedMNIST V2 images are preprocessed and tokenized with patch sizes $P \in \{28, 14, 7, 4, 2, 1\}$, then used to fine-tune an ImageNet-pretrained ViT-Small (2D: new patch embedding and classification head; 3D: weight inflation and interpolated positional embeddings). Each model is evaluated individually, and the models trained with $P = 1, 2, 4$ are additionally combined using four fusion strategies. Additional experiments examine higher resolutions, other transformer architectures, and reduced training data. The flowchart was created with the assistance of Claude (Opus 5.5).}
    \label{fig:pipeline}
\end{figure}

An overview of the proposed pipeline is shown in Figure~\ref{fig:pipeline}, and a detailed description of its components is provided in the following subsections.

\subsection{Dataset}

\begin{figure}[htbp]
    \centering
    \setlength{\tabcolsep}{-3pt}
    \begin{tabular}{c c c c}
        \subfloat[\centering \textbf{BloodMNIST}\\Microscopy | Multi (8)]{
            \includegraphics[width=0.25\textwidth]{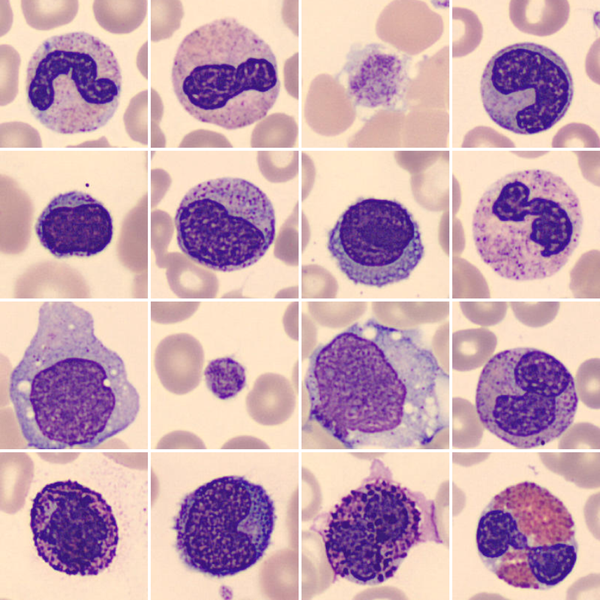}
        } &
        \subfloat[\centering \textbf{BreastMNIST}\\Ultrasound | Binary (2)]{
            \includegraphics[width=0.25\textwidth]{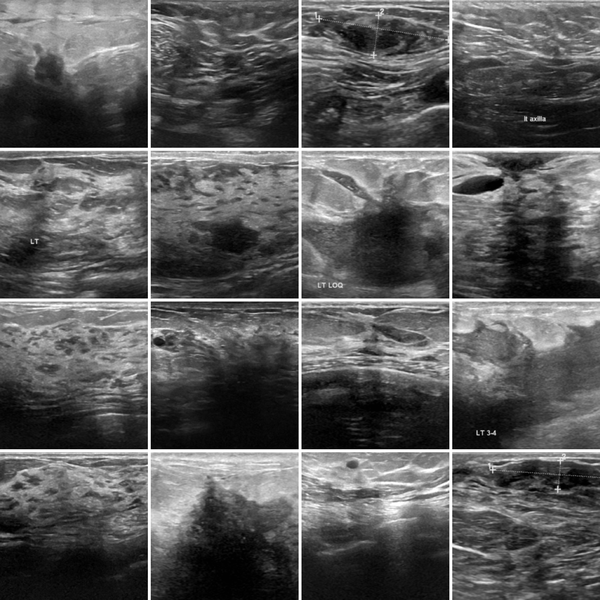}
        } &
        \subfloat[\centering \textbf{DermaMNIST}\\Dermatoscopy | Multi (7)]{
            \includegraphics[width=0.25\textwidth]{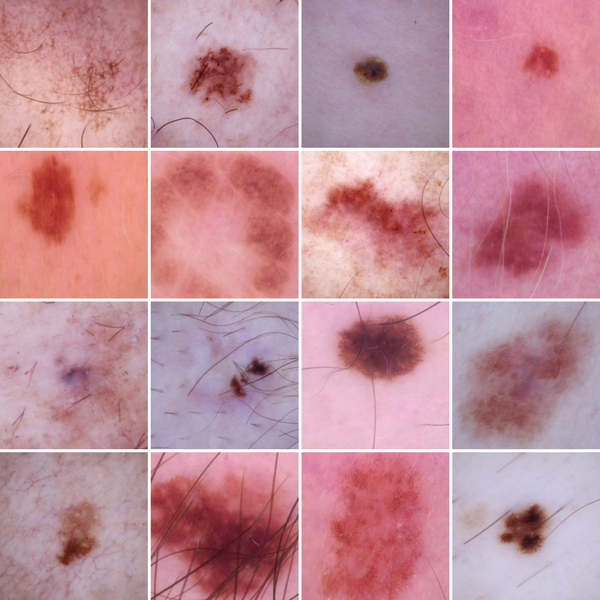}
        } &
        \subfloat[\centering \textbf{OCTMNIST}\\Retinal OCT | Multi (4)]{
            \includegraphics[width=0.25\textwidth]{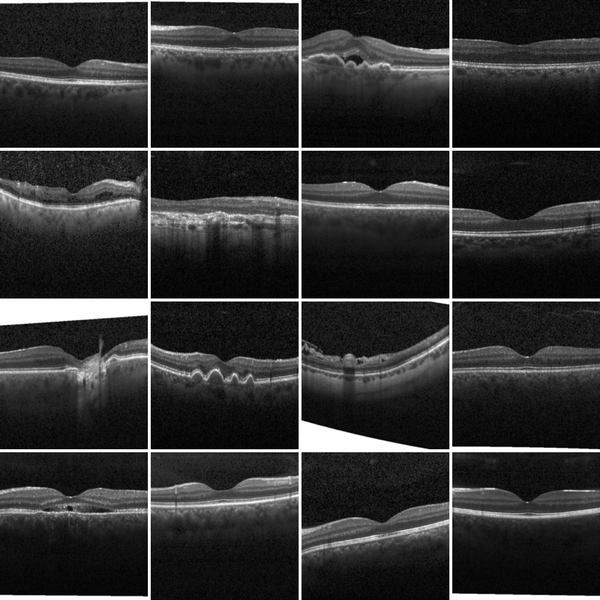}
        } \\

        \subfloat[\centering \textbf{OrganSMNIST}\\Abdominal CT | Multi (11)]{
            \includegraphics[width=0.25\textwidth]{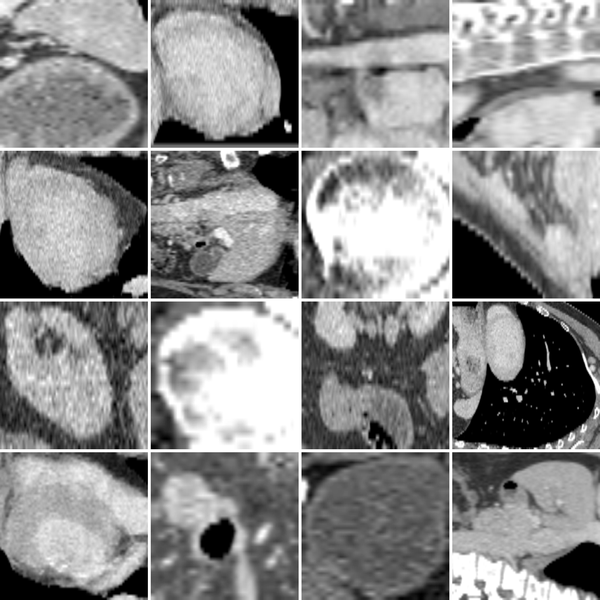}
        } &
        \subfloat[\centering \textbf{PneumoniaMNIST}\\Chest X-ray | Binary (2)]{
            \includegraphics[width=0.25\textwidth]{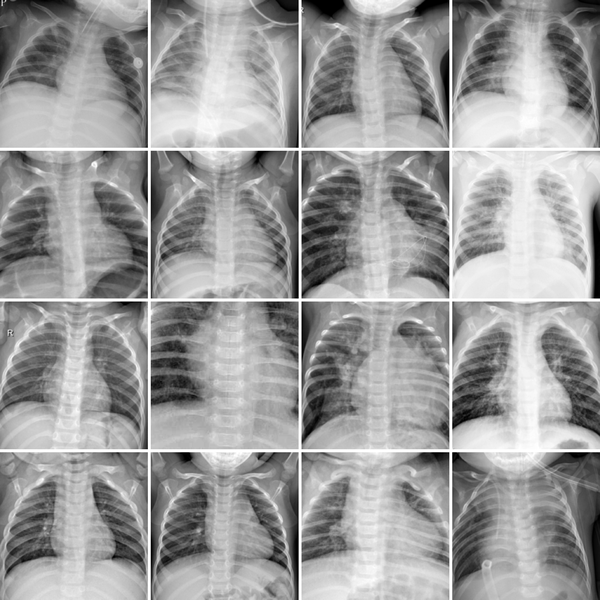}
        } &
        \subfloat[\centering \textbf{RetinaMNIST}\\Fundus imaging | Multi (5)]{
            \includegraphics[width=0.25\textwidth]{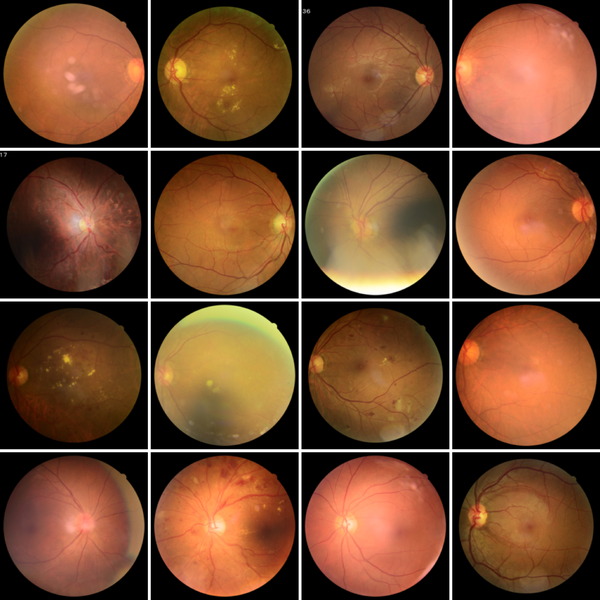}
        } &
        \subfloat[\centering \textbf{AdrenalMNIST3D}\\Abdominal CT |\\ Binary (2)]{
            \includegraphics[width=0.25\textwidth]{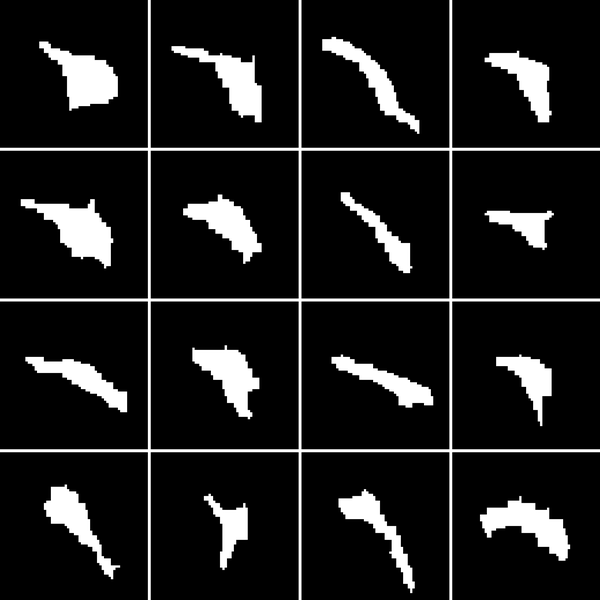}
        } \\

        \subfloat[\centering \textbf{FractureMNIST3D}\\Chest CT | Multi (3)]{
            \includegraphics[width=0.25\textwidth]{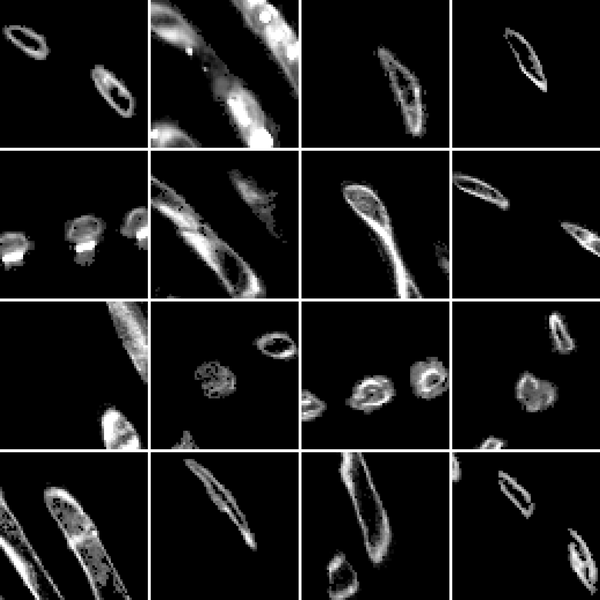}
        } &
        \subfloat[\centering \textbf{NoduleMNIST3D}\\Chest CT | Binary (2)]{
            \includegraphics[width=0.25\textwidth]{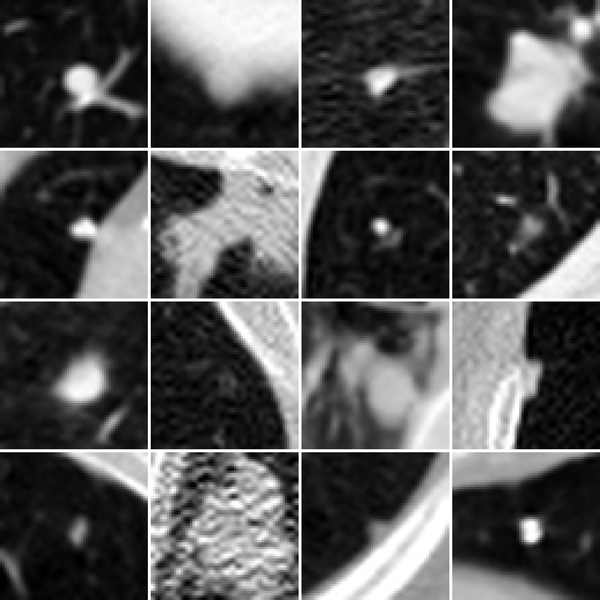}
        } &
        \subfloat[\centering \textbf{SynapseMNIST3D}\\Electron microscopy | Binary (2)]{
            \includegraphics[width=0.25\textwidth]{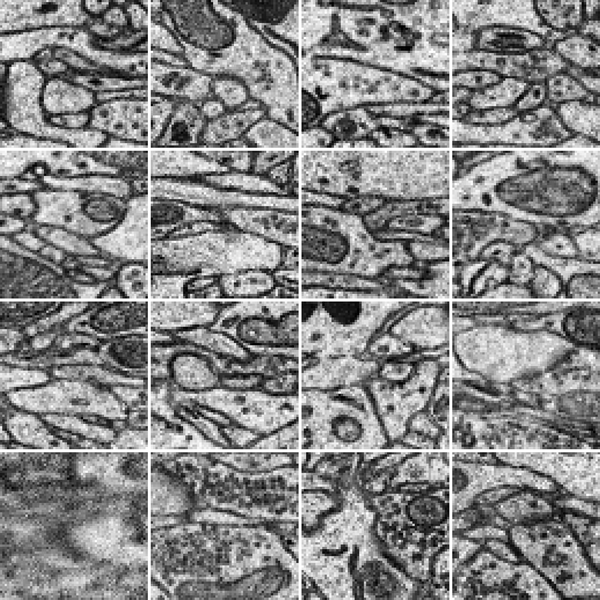}
        } &
        \subfloat[\centering \textbf{VesselMNIST3D}\\Brain MRA | Binary (2)]{
            \includegraphics[width=0.25\textwidth]{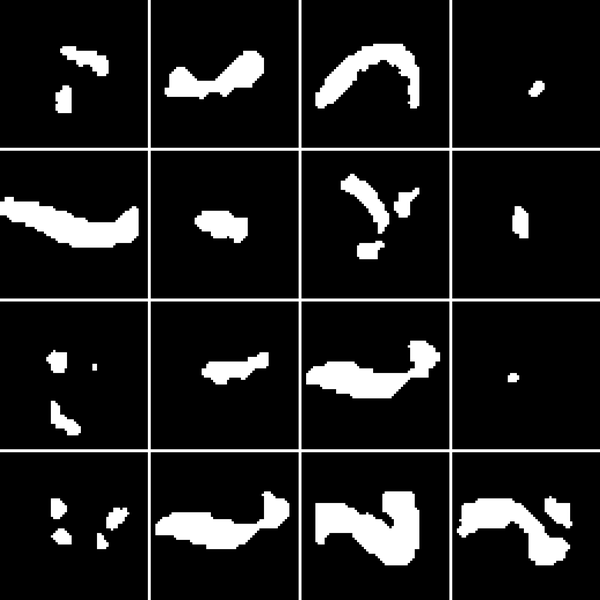}
        }
    \end{tabular}
    \caption{\centering Overview of the 12 selected datasets from the MedMNIST V2 collection used in this study. Each sub-figure shows representative example images along with the dataset name, imaging modality, and classification task. For better visibility, the examples are shown at 224$\times$224 (2D) and 64$\times$64$\times$64 (3D) resolution; for the 3D datasets, the middle slices of random samples are shown.}
    \label{fig:medmnist_overview}
\end{figure}

We use a subset of the MedMNIST V2 collection~\cite{Yang2023}, which provides a variety of biomedical image-classification datasets: seven 2D subsets and five 3D subsets from various imaging modalities, as summarized in Table~\ref{tab:medmnist}. Visual examples of the selected datasets are shown in Figure~\ref{fig:medmnist_overview}. Although MedMNIST V2 provides images at several resolutions (up to 224$\times$224 for 2D and up to 64$\times$64$\times$64 for 3D datasets), we use the smallest available size (28$\times$28 for 2D and 28$\times$28$\times$28 for 3D datasets) in our main experiments, so that all experiments can be run on a single GPU. To validate our main findings, we additionally evaluate higher-resolution images (224$\times$224 for 2D and 64$\times$64$\times$64 for 3D datasets) in a set of additional experiments.


We explicitly choose the MedMNIST V2 collection because it contains images from a wide range of imaging modalities in both 2D and 3D formats, which allows us to perform a comprehensive analysis across diverse types of medical images. The selected datasets include both binary and multi-class classification tasks, as indicated in Table~\ref{tab:medmnist}.

\begin{table}[htbp]

\centering
\caption{Summary of the selected datasets from MedMNIST V2 collection~\cite{Yang2023}.}
\begin{tabular}{lcccc}
\hline
\textbf{Dataset} & \textbf{Type} & \textbf{\# train/val/test} & \textbf{classes} & \textbf{Modality} \\
\hline
Breast & 2D & 546/78/156 & 2 & Ultrasound \\
Retina & 2D & 1080/120/400 & 5 & Fundus imaging \\
Blood & 2D & 11959/1712/3421 & 8 & Microscopy \\
Derma & 2D & 7007/1003/2005 & 7 & Dermatoscopy \\
OCT & 2D & 97477/10832/1000 & 4 & Retinal OCT \\
OrganS & 2D & 13932/2452/8827 & 11 & Abdominal CT \\
Pneumonia & 2D & 4708/524/624 & 2 & X-ray \\
\hline
Adrenal & 3D & 1188/98/298 & 2 & Abdominal CT \\
Fracture & 3D & 1027/103/240 & 3 & Chest CT \\
Nodule & 3D & 1158/165/310 & 2 &  Chest CT\\
Synapse & 3D & 1230/177/352 & 2 & Electron Microscopy \\
Vessel & 3D & 1335/191/382 & 2 & Brain MRA \\
\hline
\end{tabular}
\label{tab:medmnist}
\end{table}

\subsection{Preprocessing}
From the seven 2D datasets, five are already provided in RGB format, while two (BreastMNIST and PneumoniaMNIST) are grayscale. For the grayscale datasets, we replicate the single channel three times to create an RGB-like three-channel representation, ensuring compatibility with models that require RGB inputs. Data augmentation is applied during training to artificially increase the dataset size, including random resized cropping, horizontal flipping, rotation, and light color adjustments.

For the 3D datasets, the volumetric images (H$\times$W$\times$D) are first converted to tensors and then replicated across three channels (H$\times$W$\times$D$\times$3) to meet the model input requirements. We also apply 3D spatial augmentations, such as random flipping and rotations, during training.

\subsection{Fine-Tuning Pretrained ViT Models}

ViTs are available in multiple scales, defined by the number of transformer layers, embedding dimension, and number of attention heads. These configurations result in several variants such as Tiny, Small, Base, Large, and Huge. For our main experiments, we use the ViT-Small model, which includes 12 transformer layers, an embedding dimension of 384, and 6 attention heads, totaling approximately 22 million parameters. This configuration provides a strong balance between performance and computational efficiency. We use the ImageNet-pretrained weights~\cite{ref22}, which allow for efficient transfer learning with limited medical data. To validate our findings across different ViT-based architectures, we conduct additional experiments with ViT-Tiny, Swin Transformer (Tiny and Small)~\cite{ref12}, and FlexiViT (Small and Base)~\cite{10205121}. FlexiViT benefits from training with different patch sizes through flexible patch embeddings, whereas Swin Transformer employs a hierarchical architecture with window-based self-attention.

In 2D datasets, we replace the pretrained classification head with a new one that matches the number of classes in the target dataset, while keeping the transformer encoder trainable so it can adapt its learned features to the medical images. For ViT-Small, the patch embedding layer is set according to the selected patch size, while the pretrained transformer weights are retained. For FlexiViT, the pretrained patch embedding weights are resized to the target patch size using pseudoinverse-based resizing (PI-resize) to better preserve the pretrained information across patch sizes~\cite{10205121}. For Swin Transformer, the hierarchical stages and attention window size are tuned according to the resulting patch grid, and compatible pretrained weights are transferred to the modified architecture.

In 3D datasets, we expand the pretrained weights using the weight inflation method introduced in~\cite{ref23}. In ViT-Small, we convert the original 2D patch embeddings into 3D kernels by repeating the weights along the depth axis. This method allows us to tokenize 3D images into non-overlapping 3D cubes. The number of tokens increases compared to the 2D case because of the extra depth dimension. To maintain geometric coherence, we reshape the pretrained 2D positional embeddings into a spatial grid and interpolate them to fit the 3D token arrangement using trilinear interpolation. FlexiViT uses a similar inflation procedure for fixed-patch experiments, with an additional option for dynamically adapting patch embeddings through PI-resizing. For Swin Transformer, the 2D patch embedding kernels are resized when necessary and inflated into 3D cubic kernels. The pretrained attention and feed-forward weights are transferred to the corresponding 3D blocks, while the relative positional biases are extended to three dimensions. The hierarchical structure and attention window size are also adjusted to suit the volumetric patch grid.

\subsection{Implementation Details and Prediction Fusion}

\begin{figure}[htbp]
    \centering
    \includegraphics[width=\textwidth]{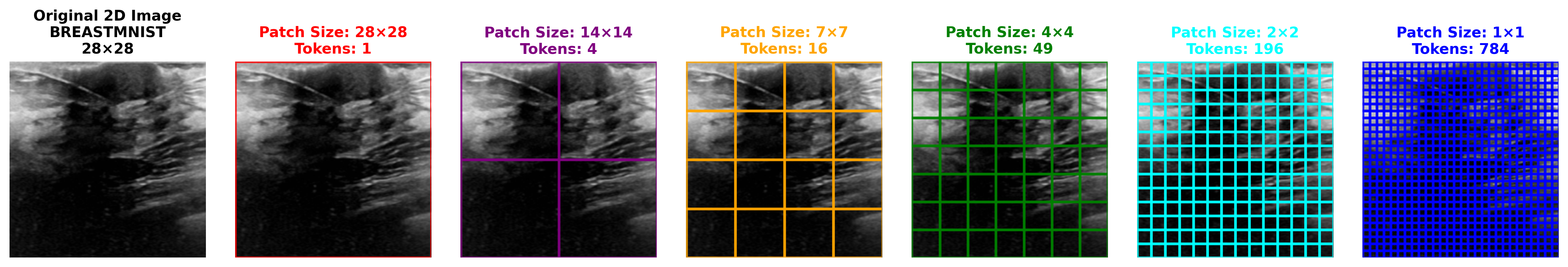}
    \caption{Example of splitting the image into patches with different sizes for 2D models.}
    \label{fig:patch_token_2d}
\end{figure}

\begin{figure}[htbp]
    \centering
    \includegraphics[width=0.9\textwidth]{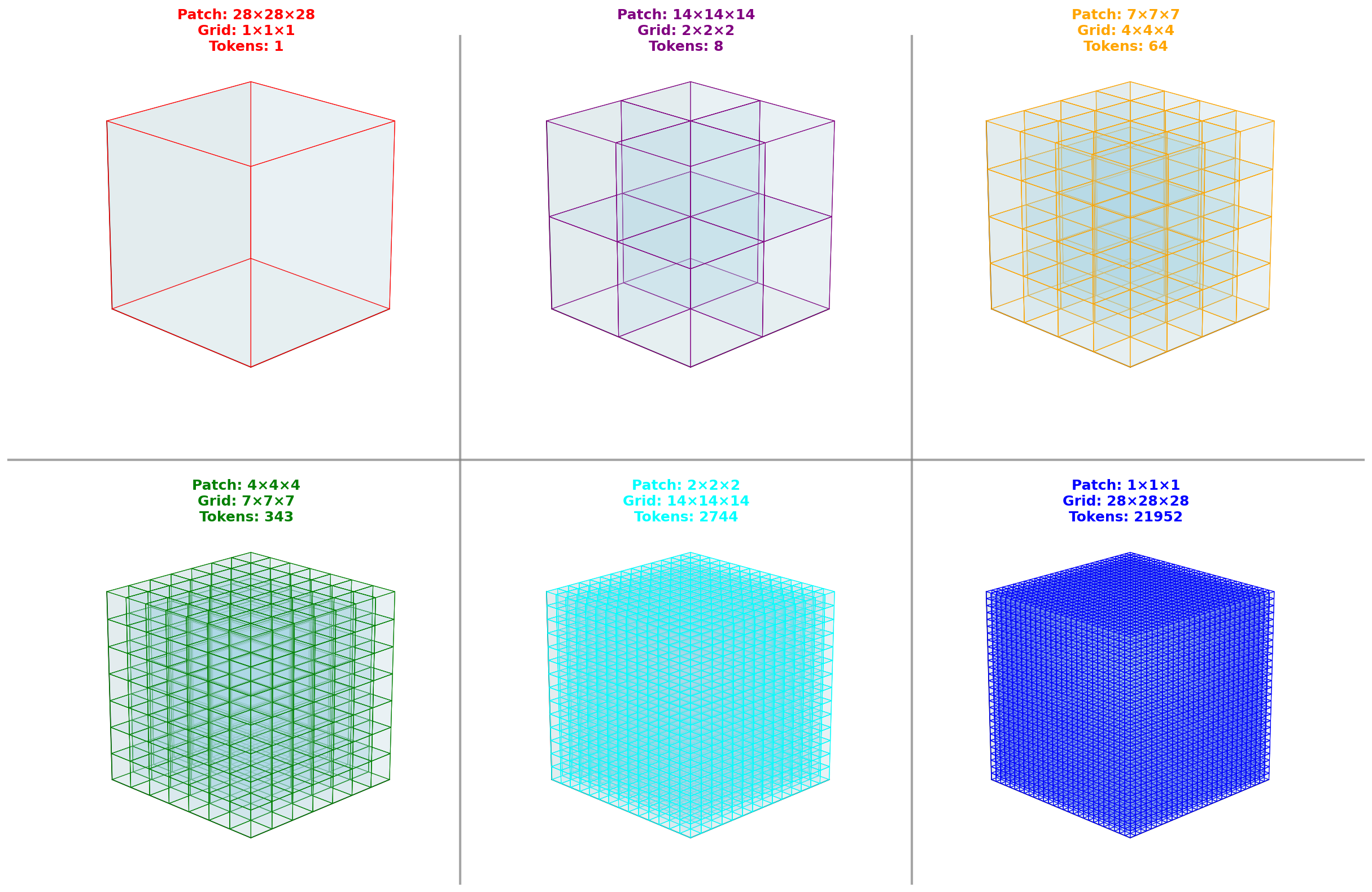}
    \caption{Example of splitting the image into patches with different sizes for 3D models.}
    \label{fig:patch_token_3d}
\end{figure}

In our main experiments, where all input images are 28$\times$28 (2D) or 28$\times$28$\times$28 (3D), we evaluate models with patch sizes $P \in \{28, 14, 7, 4, 2, 1\}$. For the additional higher-resolution experiments, we use
$P \in \{224,\allowbreak 112,\allowbreak 56,\allowbreak 32,\allowbreak 28,\allowbreak 16,\allowbreak 14,\allowbreak 8,\allowbreak 7,\allowbreak 4,\allowbreak 2,\allowbreak 1\}$ for 224$\times$224 2D images and $P \in \{64, 32, 16, 8, 4, 2, 1\}$ for 64$\times$64$\times$64 3D volumes. In all cases, the patch sizes are factors of the input resolution, which allows the image to be divided into non-overlapping patches without padding or truncation. This setup keeps spatial integrity, prevents partial patches at the edges, and makes it easier to reshape the image grid into the sequence of tokens used by the Transformer. Given an image with size of 28, $P = 28$ represents a global image-level tokenization with a single 1$\times$1 grid. $P = 14$ and $P = 7$ provide 2$\times$2 and 4$\times$4 grids for mid-level representation, respectively. Smaller patches $(P\leq4)$ capture fine local structures (see Figure~\ref{fig:patch_token_2d} and Figure~\ref{fig:patch_token_3d} for examples). This design gradually increases the number of tokens entered into the model and enables us to study how patch size affects both the model performance and computational cost.

Previous studies have shown that combining multiple sources of information, either by ensembling the predictions of multiple models or by fusing information from different modalities or feature representations, often yields better classification performance than individual models~\cite{mahbod2021pollen, 11515331, Steyaert2023, 10.1093/bioadv/vbag131}.

Motivated by these findings, we also investigate, as a complementary analysis, whether combining models trained with different small patch sizes further improves performance. Specifically, we fuse three independently fine-tuned models with patch sizes of 1, 2, and 4 using four strategies. Averaging $(A(1,2,4))$ computes the mean of the predicted class probabilities of the individual models and has no trainable parameters. Early fusion $(E(1,2,4))$ concatenates the final-layer token embeddings of the individual models and passes the result through a learnable projection layer and classification head. Cross-model attention fusion $(C(1,2,4))$ projects these embeddings into a common embedding space and applies multi-head self-attention across the resulting model tokens, followed by mean pooling and a classification head. Late fusion $(L(1,2,4))$ combines the predicted class probabilities using learnable branch weights that are normalized with a softmax, so that the weights are positive and sum to one. For the three trainable schemes, the fine-tuned backbone models are kept frozen, and only the fusion-specific parameters are trained.

All subset datasets follow standard splits into training, validation, and test partitions as provided by MedMNIST V2~\cite{Yang2023}. This supports reproducible comparisons and consistent baselines across subsets. We created two training pipelines: one for 2D image classification and another for 3D volumetric data. They have the same experimental structure, optimization routine, and evaluation protocol, while considering the architectural differences between the models. Gradients are calculated through backpropagation, and the model parameters are updated using the AdamW optimizer~\cite{loshchilov2017decoupled} with a learning rate of $10^{-4}$ and a scheduler that halves the learning rate after every 25 epochs. The models are trained for 80 epochs using the cross-entropy loss function, and the best-performing checkpoint is selected based on the lowest validation loss. Model performance on the test set is evaluated using well-known metrics, including accuracy (Acc.), balanced accuracy (Bal. Acc.), and the area under the ROC curve (AUC). For AUC computation on multi-class datasets, a one-vs-rest approach is used. Since several of the datasets are class-imbalanced, we consider Bal. Acc. as the main evaluation metric, as it averages the recall across classes and is therefore not dominated by the majority classes. All experiments, including fusions, are repeated three times to assess robustness, and results are reported as the mean $\pm$ standard deviation for all evaluation metrics.
We measure the computational complexity by computing the number of giga floating-point operations (GFLOPs) needed to perform a single forward pass during evaluation mode (batch size = 1). All experiments are conducted using a workstation with an AMD Ryzen 9 7950X CPU, 96 GB of RAM, and a single NVIDIA RTX 4090 GPU.  Our implementation is publicly available on GitHub:~\url{https://github.com/HealMaDe/MedViT}.



\section{Results and Discussion}

\label{sec:results}
The results of the main experiments, conducted on 28$\times$28 (2D) and 28$\times$28$\times$28 (3D) inputs, are summarized in terms of Bal. Acc., our main evaluation metric, in Table~\ref{tab:results_2d} for the 2D datasets and Table~\ref{tab:results_3d} for the 3D datasets. The complete results for each dataset, including Acc., Bal. Acc., AUC, and GFLOPs, are provided in Tables~S1-S12 of the supplementary material. We organize our analysis into three groups: small-scale 2D datasets (BreastMNIST and RetinaMNIST), medium- to large-scale 2D datasets (BloodMNIST, DermaMNIST, OCTMNIST, OrganSMNIST, and PneumoniaMNIST), and 3D datasets (AdrenalMNIST3D, FractureMNIST3D, NoduleMNIST3D, SynapseMNIST3D, and VesselMNIST3D). In both tables, the first six columns report the results of the individual models trained with different patch sizes, and the last four columns report the results of the fusion strategies ($A$, $E$, $C$, and $L$).

\subsection{2D Datasets}
 
The results for the small-scale 2D datasets are shown in the first two rows of Table~\ref{tab:results_2d} (see Tables~S1 and~S2 for the complete results). We discuss these results separately because fine-tuning on BreastMNIST and RetinaMNIST is affected by the limited number of training samples compared to the other 2D datasets (546 for BreastMNIST and 1080 for RetinaMNIST versus at least 4,708 training samples for the remaining 2D datasets). As the results indicate, the performance across patch sizes shows no clear trend compared with the larger datasets, suggesting that limited data may prevent the model from learning effective, patch-size-dependent features, leading to inconsistent performance patterns.
 
For BreastMNIST, averaging the predictions ($A(1,2,4)$) yields the best Bal. Acc., whereas the individual patch sizes show no clear trend. For RetinaMNIST, the best Bal. Acc. among the individual models is obtained with the largest patch size ($P = 28$), and the best overall Bal. Acc. with early fusion ($E(1,2,4)$). These results suggest that, when fine-tuning on small-scale datasets, the relationship between patch size and performance is inconsistent and does not follow the trend observed for the larger datasets (rest of the table).
 
For the medium- and large-scale 2D datasets (with training-set sizes ranging from 4,708 to 97,477), including BloodMNIST, DermaMNIST, OCTMNIST, OrganSMNIST, and PneumoniaMNIST, the impact of patch size becomes clearer and more consistent. In nearly all datasets, reducing the patch size improves classification performance. More specifically, the best performances are achieved with patch sizes of 1, 2, or 4, whereas the worst performances are consistently obtained with patch size 28. This suggests that finer spatial tokenization enables ViTs to capture local shapes and textures more effectively.
 
Among the small patch sizes (1, 2, and 4), patch size 2 achieves the best overall performance, whereas patch size 28 yields the worst performance across these datasets (average Bal. Acc. of 79.33\% for patch size 2 versus 70.31\% for patch size 28). A detailed comparison between patch sizes 2 and 28 is provided in the supplementary material (Table S13), where the largest improvement in Bal. Acc. is observed for the OrganSMNIST dataset (12.78 percentage points).
 
Furthermore, fusing the models trained with patch sizes 1, 2, and 4 further improves performance: early fusion, cross-model attention fusion, and late fusion outperform the best individual patch size in terms of Bal. Acc. on all five datasets, and averaging on four of them. However, no single fusion strategy is consistently superior, as the highest Bal. Acc. is achieved by early fusion for two datasets (BloodMNIST and DermaMNIST), by cross-model attention fusion for two datasets (OrganSMNIST and PneumoniaMNIST), and by averaging for one dataset (OCTMNIST). These results suggest that combining representations obtained at multiple tokenization scales can enhance classification performance.

\begin{table}[htbp]
\centering
\caption{Balanced accuracy (\%, mean over three runs) on the 2D datasets (28$\times$28) for different patch sizes and fusion strategies. Fusion strategies combine the models trained with patch sizes 1, 2, and 4: $A$, averaging; $E$, early fusion; $C$, cross-model attention fusion; $L$, late fusion. For each dataset, the best individual patch size and the best fusion strategy are shown in bold. GFLOPs are identical for all 2D datasets. Standard deviations and the full results, including Acc.\ and AUC, are provided in Tables~S1-S7 of the supplementary material.}
\label{tab:results_2d}
\setlength{\tabcolsep}{4pt}
\renewcommand{\arraystretch}{1.15}
\begin{adjustbox}{max width=\textwidth}
\begin{tabular}{lcccccc@{\hspace{10pt}}cccc}
\toprule
 & \multicolumn{6}{c}{\textbf{Patch size $P$}} & \multicolumn{4}{c}{\textbf{Fusion of $P = 1, 2, 4$}} \\
\cmidrule(lr){2-7}\cmidrule(lr){8-11}
\textbf{Dataset} & \textbf{1} & \textbf{2} & \textbf{4} & \textbf{7} & \textbf{14} & \textbf{28} & \textbf{A} & \textbf{E} & \textbf{C} & \textbf{L} \\
\midrule
\multicolumn{11}{l}{\textit{Small-scale datasets}} \\
BreastMNIST & 77.05 & \textbf{77.24} & 76.94 & 75.63 & 76.21 & 75.36 & \textbf{80.74} & 75.58 & 76.73 & 77.38 \\
RetinaMNIST & 36.71 & 32.02 & 34.51 & 36.01 & 35.58 & \textbf{36.99} & 34.71 & \textbf{39.66} & 38.25 & 36.85 \\
\midrule
\multicolumn{11}{l}{\textit{Medium- to large-scale datasets}} \\
BloodMNIST & 96.03 & \textbf{96.56} & 96.09 & 95.95 & 94.33 & 92.36 & 97.10 & \textbf{97.44} & 97.06 & 96.69 \\
DermaMNIST & 56.16 & 54.86 & \textbf{56.64} & 54.38 & 48.76 & 42.23 & 59.32 & \textbf{64.11} & 63.43 & 57.69 \\
OCTMNIST & 76.60 & \textbf{79.10} & 78.63 & 78.03 & 71.70 & 67.37 & \textbf{79.90} & 79.80 & 79.67 & 79.33 \\
OrganSMNIST & 74.96 & \textbf{76.51} & 76.35 & 74.42 & 70.71 & 63.73 & 78.77 & 78.97 & \textbf{79.02} & 76.87 \\
PneumoniaMNIST & 87.27 & \textbf{89.62} & 87.45 & 86.25 & 87.66 & 85.84 & 89.07 & 89.89 & \textbf{90.37} & 89.66 \\
\midrule
GFLOPs & 16.71 & 4.19 & 1.06 & 0.36 & 0.11 & 0.04 & 21.96 & 21.96 & 21.96 & 21.96 \\
\bottomrule
\end{tabular}
\end{adjustbox}
\end{table}

\subsection{3D Datasets}
 
Across the 3D datasets (AdrenalMNIST3D, FractureMNIST3D, NoduleMNIST3D, SynapseMNIST3D, and VesselMNIST3D), the results in Table~\ref{tab:results_3d} and Tables~S8-S12 show a clear trend regarding the effect of patch size on ViT performance. Similar to the 2D experiments, smaller volumetric patches generally lead to better classification results. Patch size 1 achieves the best Bal. Acc. on four of the five datasets and the second-best on the remaining one (NoduleMNIST3D, where patch size 2 performs best). As the patch size increases, the Bal. Acc. averaged across the five 3D datasets decreases steadily from 66.96\% for patch size 1 to 63.98\%, 61.18\%, and 56.61\% for patch sizes 2, 4, and 7, respectively, and remains at a similarly low level for patch sizes 14 and 28 (55.73\% and 56.40\%; see also Figure~\ref{fig:average}). Larger patch sizes therefore deliver inferior results by a large margin.

A detailed comparison between patch size 1 and patch size 14 is provided in Table S14 in the supplementary material, where the largest improvement in Bal. Acc. is observed for the VesselMNIST3D dataset (23.78 percentage points).

Additionally, the effect of fusion on the 3D datasets depends on the fusion strategy. Early fusion improves Bal. Acc. over the best individual patch size on all five 3D datasets, and cross-model attention fusion on four of them, whereas averaging and late fusion do not outperform the best individual model in terms of Bal. Acc. on any 3D dataset. Among the fusion strategies, early fusion achieves the highest Bal. Acc. on four datasets, and cross-model attention fusion on the remaining one (NoduleMNIST3D).

\begin{table}[htbp]
\centering
\caption{Balanced accuracy (\%, mean over three runs) on the 3D datasets (28$\times$28$\times$28) for different patch sizes and fusion strategies. Fusion strategies combine the models trained with patch sizes 1, 2, and 4: $A$, averaging; $E$, early fusion; $C$, cross-model attention fusion; $L$, late fusion. For each dataset, the best individual patch size and the best fusion strategy are shown in bold. GFLOPs are identical for all 3D datasets. Standard deviations and the full results, including Acc.\ and AUC, are provided in Tables~S8-S12 of the supplementary material.}
\label{tab:results_3d}
\setlength{\tabcolsep}{4pt}
\renewcommand{\arraystretch}{1.15}
\begin{adjustbox}{max width=\textwidth}
\begin{tabular}{lcccccc@{\hspace{10pt}}cccc}
\toprule
 & \multicolumn{6}{c}{\textbf{Patch size $P$}} & \multicolumn{4}{c}{\textbf{Fusion of $P = 1, 2, 4$}} \\
\cmidrule(lr){2-7}\cmidrule(lr){8-11}
\textbf{Dataset} & \textbf{1} & \textbf{2} & \textbf{4} & \textbf{7} & \textbf{14} & \textbf{28} & \textbf{A} & \textbf{E} & \textbf{C} & \textbf{L} \\
\midrule
AdrenalMNIST3D & \textbf{68.72} & 65.76 & 61.12 & 57.74 & 60.59 & 58.68 & 64.70 & \textbf{70.92} & 68.56 & 63.85 \\
FractureMNIST3D & \textbf{52.56} & 49.51 & 47.79 & 44.51 & 40.05 & 42.34 & 51.48 & \textbf{55.34} & 54.99 & 51.69 \\
NoduleMNIST3D & 76.91 & \textbf{78.50} & 75.23 & 72.69 & 72.11 & 72.85 & 76.86 & 78.88 & \textbf{79.17} & 76.91 \\
SynapseMNIST3D & \textbf{56.93} & 53.42 & 50.80 & 51.05 & 50.00 & 50.27 & 52.02 & \textbf{59.94} & 58.48 & 52.02 \\
VesselMNIST3D & \textbf{79.67} & 72.70 & 70.95 & 57.05 & 55.89 & 57.88 & 73.29 & \textbf{83.50} & 82.00 & 73.24 \\
\midrule
GFLOPs & 102.12 & 15.03 & 2.54 & 0.39 & 0.07 & 0.05 & 119.69 & 119.69 & 119.69 & 119.69 \\
\bottomrule
\end{tabular}
\end{adjustbox}
\end{table}

\subsection{Additional Experiments}
 
To further examine the generalizability of our findings, we conduct three additional experiments on OrganSMNIST and VesselMNIST3D, as these datasets show the largest differences in Bal. Acc. between small and large patch sizes among the 2D and 3D datasets, respectively (Tables~S13 and~S14). Consistent with the previous sections, we report Bal. Acc. in the main text (Tables~\ref{tab:add_resolution}-\ref{tab:add_reduced}), while the complete results, including Acc., AUC, and standard deviations, are provided in Tables~S15-S20 of the supplementary material.
 
First, we investigate whether the performance improvements observed with smaller patch sizes at the original resolution of 28 are retained at higher image resolutions. For this purpose, we use OrganSMNIST at 224$\times$224 and VesselMNIST3D at 64$\times$64$\times$64, which are the highest resolutions available in the MedMNIST V2 collection. To make these experiments feasible on a single GPU, we use ViT-Tiny instead of the ViT-Small model used in the main experiments. The results are presented in Table~\ref{tab:add_resolution}.

For OrganSMNIST, Bal. Acc. improves substantially as the patch size decreases from 224 (50.05\%) and 112 (61.74\%) to 56 (76.37\%), remains at a similar level for patch sizes between 7 and 56 (75.52\% to 77.21\%), and reaches its highest value with patch size 4 (77.77\%). Further reducing the patch size to 2 and 1 lowers Bal. Acc. to 74.09\% and 71.99\%, respectively, indicating that the smallest patch sizes are not optimal at this resolution. This pattern is consistent with the results at 28$\times$28, where the best Bal. Acc. is likewise obtained with a small but not the smallest patch size, both for ViT-Small in the main experiments ($P = 2$, 76.51\%, followed by 74.96\% for $P = 1$; Table~\ref{tab:results_2d}) and for ViT-Tiny, as presented later in this section ($P = 2$, 75.58\%, followed by 73.42\% for $P = 1$; Table~\ref{tab:add_architectures}). For VesselMNIST3D, Bal. Acc. improves consistently as the patch size decreases, from 49.51\% with patch size 64 to 81.02\% with patch size 1. This is also in line with the results at 28$\times$28$\times$28, where both ViT-Small (79.67\%; Table~\ref{tab:results_3d}) and ViT-Tiny (77.83\%; Table~\ref{tab:add_architectures}) achieve their best Bal. Acc. with patch size 1. Compared with ViT-Tiny at the original resolution, the higher resolution leads to a moderate improvement in the best achievable Bal. Acc. (77.77\% vs.\ 75.58\% for OrganSMNIST and 81.02\% vs.\ 77.83\% for VesselMNIST3D). However, these improvements come at a substantial computational cost, as GFLOPs increase from 0.04 to 267.59 for OrganSMNIST and from 0.02 to 842.06 for VesselMNIST3D when the patch size is reduced from the largest to the smallest value.

\begin{table}[htbp]
\centering
\caption{Balanced accuracy (\%, mean over three runs) and computational cost (GFLOPs) of ViT-Tiny for different patch sizes at higher image resolutions. The best result for each dataset is shown in bold. Standard deviations and the full results, including Acc.\ and AUC, are provided in Tables~S15 and~S16 of the supplementary material.}
\label{tab:add_resolution}
\setlength{\tabcolsep}{6pt}
\renewcommand{\arraystretch}{1.1}
\begin{adjustbox}{max width=\textwidth}
\begin{tabular}[t]{ccc}
\toprule
\multicolumn{3}{c}{\textbf{OrganSMNIST}} \\
\multicolumn{3}{c}{(224$\times$224)} \\
\midrule
\textbf{Patch size $P$} & \textbf{Bal. Acc.} & \textbf{GFLOPs} \\
\midrule
1 & 71.99 & 267.59 \\
2 & 74.09 & 66.92 \\
4 & \textbf{77.77} & 16.76 \\
7 & 75.52 & 5.49 \\
8 & 75.84 & 4.21 \\
14 & 75.72 & 1.40 \\
16 & 77.01 & 1.08 \\
28 & 77.10 & 0.38 \\
32 & 77.21 & 0.30 \\
56 & 76.37 & 0.12 \\
112 & 61.74 & 0.06 \\
224 & 50.05 & 0.04 \\
\bottomrule
\end{tabular}
\hspace{20pt}
\begin{tabular}[t]{ccc}
\toprule
\multicolumn{3}{c}{\textbf{VesselMNIST3D}} \\
\multicolumn{3}{c}{(64$\times$64$\times$64)} \\
\midrule
\textbf{Patch size $P$} & \textbf{Bal. Acc.} & \textbf{GFLOPs} \\
\midrule
1 & \textbf{81.02} & 842.06 \\
2 & 72.65 & 112.55 \\
4 & 61.65 & 16.02 \\
8 & 61.06 & 2.56 \\
16 & 54.01 & 0.50 \\
32 & 51.21 & 0.09 \\
64 & 49.51 & 0.02 \\
\bottomrule
\end{tabular}
\end{adjustbox}
\end{table}
 
In another experiment, we investigate whether the observed performance improvements from reducing the patch size generalize across different transformer architectures. For this purpose, we evaluate six models: ViT-Tiny and ViT-Small, Swin-Tiny and Swin-Small, and FlexiViT-Small and FlexiViT-Base. These architectures differ primarily in their attention mechanisms and patch embedding strategies. 
The results are presented in Table~\ref{tab:add_architectures}.
 
For OrganSMNIST, all six architectures achieve their best Bal. Acc. with a patch size of 4 or smaller and their worst with patch size 28, although the optimal patch size varies across architectures. ViT-Tiny and ViT-Small perform best with patch size 2, both Swin variants with patch size 1, and both FlexiViT variants with patch size 4 (for FlexiViT-Base, equally with patch size 7). Among all evaluated models, Swin-Small achieves the highest Bal. Acc. (77.28\%). 

For VesselMNIST3D, all six architectures achieve their best Bal. Acc. with patch size 1 or 2, and ViT-Small achieves the highest Bal. Acc. among all models (79.67\%). However, the relationship between patch size and performance is not strictly monotonic for the 3D dataset, as the lowest Bal. Acc. is obtained with intermediate patch sizes (4, 7, or 14) rather than with patch size 28 for all architectures.

Among the evaluated architectures, FlexiViT is the least sensitive to the choice of patch size on both datasets. On OrganSMNIST, the difference between the best and worst Bal. Acc. across patch sizes is 7.90 and 5.60 percentage points for FlexiViT-Small and FlexiViT-Base, respectively, compared with 11.40 to 25.24 percentage points for the other architectures. On VesselMNIST3D, this difference is also the smallest (8.28 and 11.09 percentage points, compared with 11.64 to 23.78 percentage points for the other architectures). This behavior may be explained by the design of FlexiViT, which is pretrained with randomly sampled patch sizes and is therefore intended to perform consistently across different patch sizes~\cite{10205121}. At the same time, both FlexiViT variants achieve the lowest best Bal. Acc. among all architectures on both datasets (at most 71.77\% for OrganSMNIST and 71.10\% for VesselMNIST3D). A possible reason is that the smallest patch sizes evaluated in our experiments (1 to 7) lie below the range of patch sizes used during FlexiViT pretraining (8 to 48 at a resolution of 240$\times$240), and the original authors report that FlexiViT does not generalize well to patch sizes smaller than those seen during training~\cite{10205121}. This is supported by our results, as FlexiViT performs comparably to or better than most other architectures with patch sizes 14 and 28, which lie within its pretraining range, but benefits less from smaller patch sizes.

All in all, these results suggest that the advantages of smaller patches extend to different transformer architectures in both 2D and 3D classification, although the optimal patch size depends on the architecture. FlexiViT is the least sensitive to the choice of patch size but benefits less from the smallest patch sizes than the other architectures.

\begin{table}[htbp]
\centering
\caption{Balanced accuracy (\%, mean over three runs) of different transformer architectures for different patch sizes on OrganSMNIST (28$\times$28) and VesselMNIST3D (28$\times$28$\times$28). For each architecture, the best patch size is shown in bold. Standard deviations and the full results, including Acc.\ and AUC, are provided in Tables~S17 and~S18 of the supplementary material.}
\label{tab:add_architectures}
\setlength{\tabcolsep}{6pt}
\renewcommand{\arraystretch}{1.1}
\begin{adjustbox}{max width=\textwidth}
\begin{tabular}{lcccccc}
\toprule
 & \multicolumn{6}{c}{\textbf{Patch size $P$}} \\
\cmidrule(lr){2-7}
\textbf{Architecture} & \textbf{1} & \textbf{2} & \textbf{4} & \textbf{7} & \textbf{14} & \textbf{28} \\
\midrule
\multicolumn{7}{l}{\textit{OrganSMNIST}} \\
ViT-Tiny & 73.42 & \textbf{75.58} & 74.37 & 73.64 & 69.68 & 64.18 \\
ViT-Small & 74.96 & \textbf{76.51} & 76.35 & 74.42 & 70.71 & 63.73 \\
Swin-Tiny & \textbf{76.67} & 73.52 & 64.12 & 67.16 & 62.24 & 53.35 \\
Swin-Small & \textbf{77.28} & 72.35 & 62.03 & 66.25 & 61.82 & 52.04 \\
FlexiViT-Small & 68.98 & 70.62 & \textbf{71.77} & 71.30 & 67.90 & 63.87 \\
FlexiViT-Base & 68.99 & 70.02 & \textbf{70.31} & \textbf{70.31} & 69.66 & 64.71 \\
\midrule
\multicolumn{7}{l}{\textit{VesselMNIST3D}} \\
ViT-Tiny & \textbf{77.83} & 74.68 & 66.54 & 57.19 & 57.44 & 59.33 \\
ViT-Small & \textbf{79.67} & 72.70 & 70.95 & 57.05 & 55.89 & 57.88 \\
Swin-Tiny & 73.84 & \textbf{79.07} & 70.07 & 65.56 & 65.68 & 68.27 \\
Swin-Small & \textbf{78.07} & 77.92 & 66.43 & 66.45 & 66.75 & 70.95 \\
FlexiViT-Small & 66.31 & \textbf{69.63} & 67.66 & 61.35 & 66.62 & 62.10 \\
FlexiViT-Base & 70.45 & \textbf{71.10} & 70.49 & 60.01 & 70.27 & 68.72 \\
\bottomrule
\end{tabular}
\end{adjustbox}
\end{table}

Finally, we investigate whether the benefits of reducing patch size persist when the number of training samples is reduced. For this purpose, we evaluate ViT-Small on OrganSMNIST and VesselMNIST3D using 100\%, 80\%, 60\%, 40\%, and 20\% of the original training data. The results are presented in Table~\ref{tab:add_reduced}. For OrganSMNIST, patch size 2 achieves the highest Bal. Acc. for all training-set sizes, and its advantage over patch size 28 is maintained as the amount of training data decreases (12.78 percentage points with the full training set and 14.69 percentage points with 20\% of the training data). For VesselMNIST3D, patch size 1 achieves the highest Bal. Acc. when 40\% or more of the training data are used, while patch size 2 performs best with 20\% of the training data. Here, the advantage of the best patch size over patch size 28 remains stable when 60\% or more of the training data are used (21.79 to 22.42 percentage points) but decreases with smaller training sets (15.92 and 11.85 percentage points with 40\% and 20\% of the training data, respectively), while still remaining substantial. With only 20\% of the training data, models with patch sizes of 4 or larger perform close to chance level (50.00\% to 53.67\%), whereas patch sizes 1 and 2 still achieve 62.29\% and 63.40\%, respectively. These findings indicate that smaller patches remain beneficial even when training data are reduced.

\begin{table}[htbp]
\centering
\caption{Balanced accuracy (\%, mean over three runs) of ViT-Small for different patch sizes when trained on different fractions of the original training set of OrganSMNIST (28$\times$28) and VesselMNIST3D (28$\times$28$\times$28). For each training-set size, the best patch size is shown in bold. Standard deviations and the full results, including Acc.\ and AUC, are provided in Tables~S19 and~S20 of the supplementary material.}
\label{tab:add_reduced}
\setlength{\tabcolsep}{6pt}
\renewcommand{\arraystretch}{1.1}
\begin{adjustbox}{max width=\textwidth}
\begin{tabular}{rccccc}
\toprule
 & \multicolumn{5}{c}{\textbf{Fraction of training data}} \\
\cmidrule(lr){2-6}
\textbf{Patch size $P$} & \textbf{100\%} & \textbf{80\%} & \textbf{60\%} & \textbf{40\%} & \textbf{20\%} \\
\midrule
\multicolumn{6}{l}{\textit{OrganSMNIST}} \\
1 & 74.96 & 74.01 & 73.40 & 72.44 & 68.67 \\
2 & \textbf{76.51} & \textbf{75.90} & \textbf{76.02} & \textbf{75.61} & \textbf{74.19} \\
4 & 76.35 & 75.78 & 75.47 & 74.20 & 72.26 \\
7 & 74.42 & 75.60 & 74.78 & 74.03 & 71.20 \\
14 & 70.71 & 70.41 & 69.88 & 69.58 & 66.80 \\
28 & 63.73 & 63.75 & 62.50 & 62.58 & 59.50 \\
\midrule
\multicolumn{6}{l}{\textit{VesselMNIST3D}} \\
1 & \textbf{79.67} & \textbf{79.96} & \textbf{80.06} & \textbf{72.68} & 62.29 \\
2 & 72.70 & 73.81 & 64.66 & 65.72 & \textbf{63.40} \\
4 & 70.95 & 61.45 & 53.00 & 56.14 & 50.00 \\
7 & 57.05 & 53.33 & 57.06 & 49.95 & 53.67 \\
14 & 55.89 & 53.86 & 55.66 & 53.48 & 50.00 \\
28 & 57.88 & 57.73 & 57.64 & 56.76 & 51.55 \\
\bottomrule
\end{tabular}
\end{adjustbox}
\end{table}

\subsection{Overall Findings and Computational Cost}

Our experimental results show improved classification performance when ViT models are fine-tuned with smaller patch sizes for both 2D and 3D datasets. However, this performance gain comes at the cost of increased computational complexity, as reflected in the GFLOPs reported in Tables~\ref{tab:results_2d} and~\ref{tab:results_3d}. For the 2D datasets at 28$\times$28 resolution, reducing the patch size from 28 to 1 improves performance but raises the computational cost from 0.04 GFLOPs to 16.71 GFLOPs per image for the ViT-Small architecture. For the 3D datasets at 28$\times$28$\times$28 resolution, the effect is even more pronounced. Decreasing the patch size of the ViT-Small model from 28 to 1 increases the computational cost from approximately 0.05 GFLOPs to more than 100 GFLOPs per volume. Although the computational cost increases with smaller patch sizes, it is important to note that this increase is not solely due to the overhead of self-attention. Other components such as the classification head, patch embedding, and various constant or linear terms also contribute to the FLOPs. In summary, for the 2D datasets, the GFLOPs increase by a factor of about four when the patch size is halved at small patch sizes (e.g., from 4 to 2 and from 2 to 1), while for the 3D datasets the increase between successive patch sizes is typically around a factor of five to seven. At larger patch sizes, the increase is smaller for both 2D and 3D datasets, particularly between patch sizes 14 and 28, where very few tokens remain and constant overheads begin to dominate the total computational cost. Besides the main results reported in the tables, we also provide the confusion matrices of the models trained with patch sizes 1, 2, and 4 and of the best-performing fusion strategy for each dataset from our primary ViT-Small experiments at a resolution of 28 in Figures S1-S4 in the supplementary material.



\begin{figure*}[htbp]
    \centering
    \includegraphics[width=\textwidth]{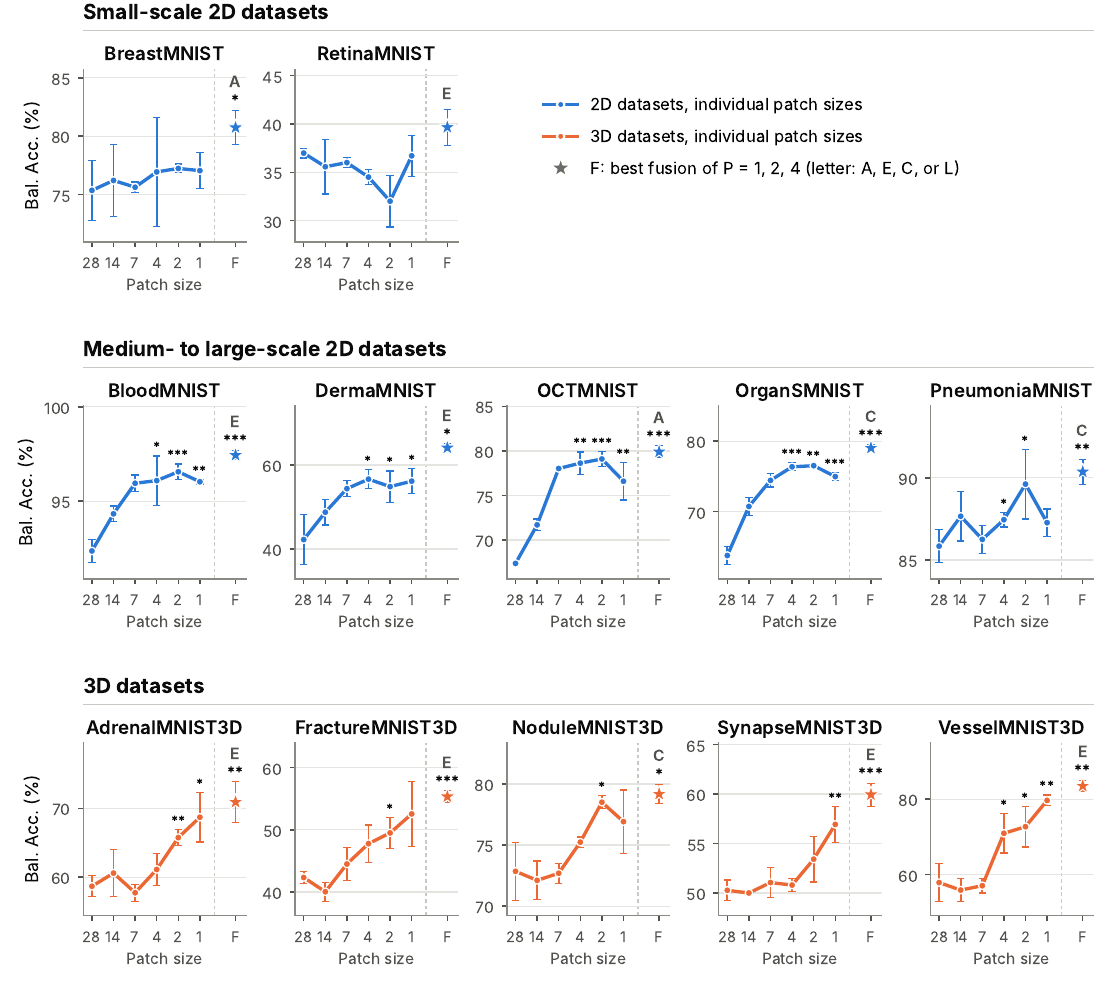}
    \caption{Balanced accuracy (\%) of ViT-Small for different patch sizes and for the best-performing fusion strategy (F) of the models trained with patch sizes 1, 2, and 4 on the small-scale 2D, medium- to large-scale 2D, and 3D datasets. Points and error bars show the mean and the standard deviation over three runs. The letter above each star indicates the best-performing fusion strategy (A: averaging, E: early fusion, C: cross-model attention, L: late fusion). Asterisks indicate a statistically significant improvement over patch size 28 based on Welch's $t$-test over the three runs (* $p < 0.05$, ** $p < 0.01$, *** $p < 0.001$).}
    \label{fig:linegraph}
\end{figure*}

For a visual summary of the main results, Figure~\ref{fig:linegraph} shows Bal. Acc. for all patch sizes and for the best-performing fusion strategy on each of the 12 datasets, grouped into small-scale 2D, medium- to large-scale 2D, and 3D datasets. To assess whether the improvements over the largest patch size are consistent across runs, we compared patch sizes 4, 2, and 1 and the best fusion strategy with patch size 28 on each dataset using Welch's $t$-test~\cite{doi:10.1177/0004563221992088} over the three runs. 
The best fusion strategy significantly outperforms patch size 28 on all datasets except RetinaMNIST, and at least one of patch sizes 4, 2, and 1 does so on all datasets except the two small-scale 2D datasets, BreastMNIST and RetinaMNIST.

\begin{figure*}[htbp]
    \centering
    \includegraphics[width=\textwidth]{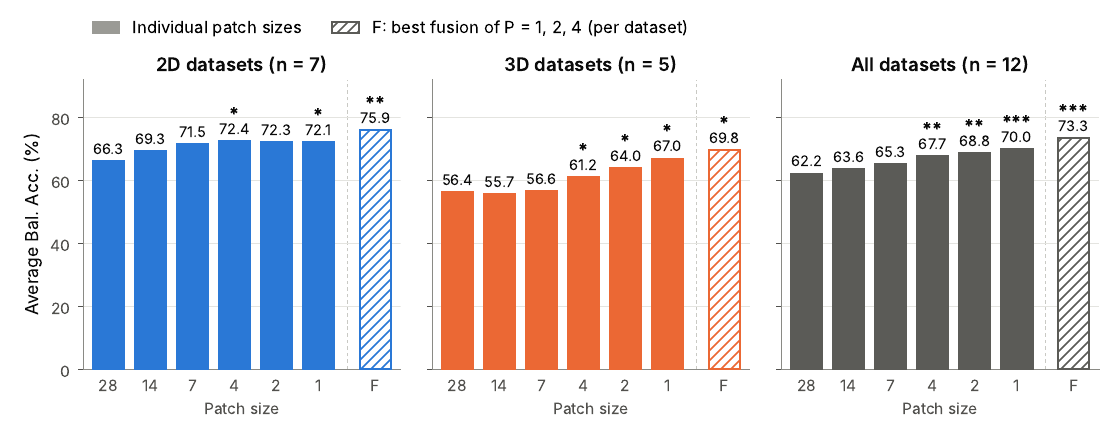}
    \caption{Average balanced accuracy (\%) of ViT-Small for different patch sizes and for the best-performing fusion strategy (F) of the models trained with patch sizes 1, 2, and 4 across all 2D datasets, all 3D datasets, and all datasets combined. For F, the best-performing fusion strategy of each dataset is used before averaging. Asterisks indicate a statistically significant improvement over patch size 28 based on the Wilcoxon signed-rank test across datasets (* $p < 0.05$, ** $p < 0.01$, *** $p < 0.001$).}
    \label{fig:average}
\end{figure*}

Figure~\ref{fig:average} shows the corresponding average performance across all 2D datasets, all 3D datasets, and all datasets combined. Overall, performance generally increases as the patch size decreases, and the best fusion strategy further improves the results on average for both 2D and 3D datasets. To assess whether these improvements generalize across datasets, we applied the Wilcoxon signed-rank test~\cite{1248547.1248548} to the mean Bal. Acc. values of the datasets, comparing each of P = 4, 2, 1, and the best fusion strategy against P = 28. Considering all 12 datasets together, patch sizes 4, 2, and 1 and the best fusion strategy achieve a significantly higher Bal. Acc. than patch size 28 ($p \leq 0.004$). For the 2D datasets alone, the improvement is significant for patch sizes 4 and 1 and for the best fusion strategy ($p \leq 0.039$). For the 3D datasets, all four comparisons reach $p = 0.031$, which is the smallest $p$-value attainable with five datasets and reflects an improvement on every 3D dataset.

Our additional experiments provide further insights into the generalizability of these findings. At higher image resolutions, smaller patches continue to offer substantial benefits, especially for 3D datasets. The comparison across vanilla ViT, Swin Transformer, and FlexiViT also suggests that the benefit of small patches is not restricted to a single architecture, although the optimal patch size varies across models. Among these architectures, FlexiViT is the least sensitive to patch size but also benefits the least from small patches, resulting in the lowest best performance on both datasets. Moreover, experiments with reduced training data also show that the benefits of small patches are generally maintained even when the amount of training data is reduced. Collectively, these findings suggest that patch size should be treated as an important hyperparameter that is selected jointly with the image resolution, model architecture, and dataset, rather than fixed to a default value.

\subsection{Comparison with CNN Baselines}
Although the main focus of this work is the impact of patch size on ViT performance, we also compare our models with the CNN-based baselines reported in the MedMNIST V2 study~\cite{Yang2023}. These baselines are based on ResNet~\cite{He2016}, one of the most widely used CNN architectures and the main benchmark model family in MedMNIST V2, namely ResNet-18 and ResNet-50 trained on 28$\times$28 and 224$\times$224 inputs for the 2D datasets, and ResNet-18 and ResNet-50 with 2.5D, 3D, and ACS convolutions~\cite{9314699} for the 3D datasets. Since these baselines are reported only in terms of AUC and Acc., we use these two metrics for the comparison. The results are presented in Tables~S21 and~S22 of the supplementary material, where ViT\_P1, ViT\_P2, and ViT\_P4 denote individual ViT models trained with patch sizes 1, 2, and 4, respectively, and ViT\_A, ViT\_E, ViT\_C, and ViT\_L denote the averaging, early, cross-model attention, and late fusion of these three models.

For all seven 2D datasets, the ViT models outperform all CNN-based baselines in terms of both AUC and Acc. With the exception of Acc. on BreastMNIST, this also holds for the best individual ViT model without fusion. Notably, ViT models outperform ResNet models trained on 224$\times$224 inputs, although they use only 28$\times$28 inputs.
 
For the 3D datasets, the comparison reveals a more heterogeneous pattern. ViT models achieve the highest AUC and Acc. on FractureMNIST3D, NoduleMNIST3D, and VesselMNIST3D, and the highest Acc. on AdrenalMNIST3D, where ResNet-18 with ACS convolutions obtains a slightly higher AUC (0.839 vs.\ 0.834). On SynapseMNIST3D, however, ResNet-50 with 3D convolutions clearly outperforms ViT models (AUC of 0.851 and Acc. of 0.795 vs.\ 0.643 and 0.727). Overall, these comparisons show that ViTs with small patch sizes are competitive with established CNN-based approaches on MedMNIST, particularly for the 2D datasets, even though our models were trained without systematic hyperparameter tuning.

\subsection{Qualitative Analysis of Attention Maps}

\begin{figure*}[htbp]
    \centering
    \includegraphics[width=0.8\textwidth]{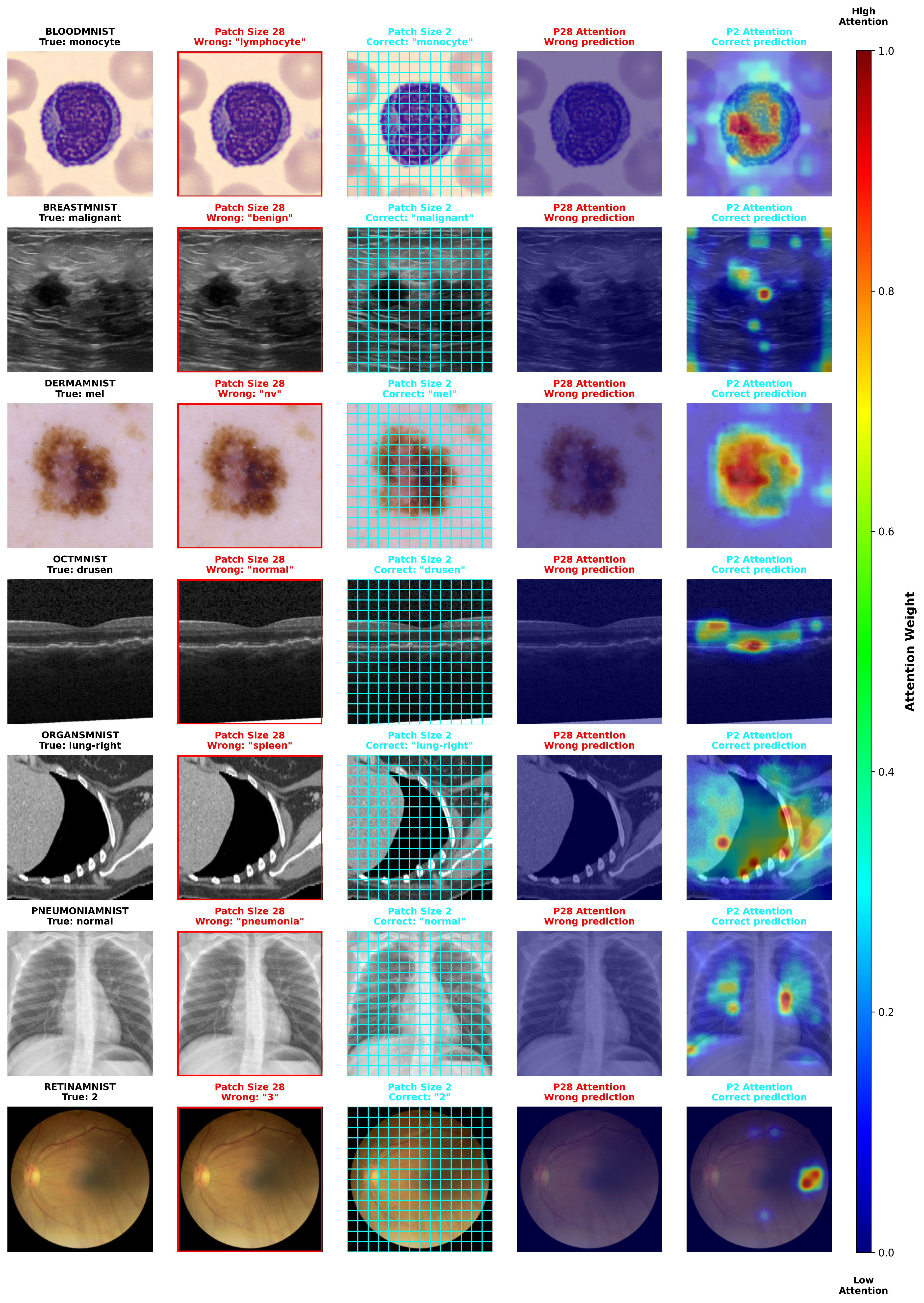}
    \caption{
    Comparison of attention maps for patch sizes of 2 and 28 for some example images. The figure shows original images with patch grids (left three columns) and corresponding attention heatmaps (right two columns). Models with smaller patch sizes (P2) consistently achieve correct predictions with more focused attention patterns, while larger patch sizes (P28) lead to incorrect predictions with less informative attention distributions.
    }
    \label{fig:patch_attention_comparison_2d}
\end{figure*}

 For a visual comparison between patch size 2 and patch size 28, we extracted attention maps from the final transformer layer for some sample images. We selected ViTs with patch sizes 2 (P2) and 28 (P28), representing the overall best and worst performing configurations. We then identified samples that were misclassified by the P28 model but correctly classified by the P2 model and present the corresponding attention heatmaps for representative examples in Figure~\ref{fig:patch_attention_comparison_2d}. As observed, models using smaller patch sizes (P2) exhibit more focused and detailed attention on diagnostically relevant regions. In contrast, models with larger patches of 28 show uniform attention patterns. This suggests that localized attention contributes to improved feature extraction for the classification task.

\subsection{Limitations} 
Despite providing an in-depth analysis of patch-size effects in ViTs for medical image classification, this study has some limitations. First, all experiments were conducted on a single GPU, which limited the choice of model capacity. As a result, our experiments were mainly based on the ViT-Small architecture, and larger variants such as ViT-Base, ViT-Large, and ViT-Huge were not evaluated. Larger models such as ViT-Large or ViT-Huge could also be considered in future analyses and might benefit even more from detailed tokenization. However, running these models would require substantially more GPU memory and computing power, which was outside the scope of this work.

Second, although we demonstrated clear benefits from using smaller patch sizes, this improvement comes with high computational costs, especially for 3D datasets. This poses challenges for deploying such models in real-world clinical settings, particularly in hospitals with limited hardware resources or scenarios requiring real-time processing.

Finally, our experiments were based on MedMNIST V2. While this allowed for a systematic comparison across many datasets from different medical imaging modalities, these datasets may not fully reflect the resolution and complexity of real clinical images. Our additional experiments at 224$\times$224 (2D) and 64$\times$64$\times$64 (3D) partially address this limitation; however, they were conducted on two datasets, and these resolutions are still considerably lower than those of many clinical images, such as full-resolution radiographs, CT and MRI volumes. Future work should therefore investigate whether the patch-size trends observed here also hold for real-world high-resolution clinical datasets.

\section{Conclusion}
\label{sec:conclusion}
In this work, we present the first thorough investigation of how patch size affects ViT classification performance in both 2D and 3D medical imaging. Using 12 datasets from the MedMNIST V2 collection, our experiments show that using smaller patch sizes leads to improved performance for both 2D and 3D datasets. This trend is particularly more evident for medium- to large-scale 2D datasets and for all 3D datasets, indicating that finer tokenization enables ViTs to capture more localized and clinically relevant features.

However, these performance gains come with substantial computational costs, as attention complexity grows rapidly with the number of tokens. This challenge is especially pronounced for 3D ViTs, where volumetric patching results in substantially larger token counts. Nevertheless, our experiments demonstrate that such analyses are still feasible on a single GPU when appropriately small image sizes are used, highlighting that patch-size studies can be performed without large-scale computational infrastructure.

Overall, our findings reveal a clear connection between patch size and model performance, providing practical guidance for designing ViT-based pipelines in medical image analysis.

\section*{Declaration of Competing Interest}
The authors declare that they have no known competing financial interests that could have appeared to influence the work reported in this paper.

\section*{Data Availability}

The MedMNIST V2 datasets used in this study are publicly available at \url{https://medmnist.com/v2}~\cite{Yang2023}. The code developed for this study, including guidelines for running it, is publicly available at \url{https://github.com/HealMaDe/MedViT}.

\section*{Ethics Statement}
Ethical approval was not required for this study, as all datasets used are publicly available and were previously published~\cite{Yang2023}.

\section*{CRediT Authorship Contribution Statement}
\textbf{Massoud Dehghan:} Conceptualization, Methodology, Writing – original draft, Writing – review and editing, Formal analysis, Software, Validation.
\textbf{Ramona Woitek:} Supervision, Validation, Writing – review and editing, Project administration.
\textbf{Amirreza Mahbod:} Supervision, Conceptualization, Methodology, Writing – original draft, Writing – review and editing, Formal analysis, Software, Validation, Project administration. 

\section*{Funding statement}
This research did not receive any specific grant from funding agencies in the public, commercial, or not-for-profit sectors.


\section*{Declaration of generative AI and AI-assisted technologies in the manuscript preparation process}

During the preparation of this work, the authors used Claude (Opus 5.5) and ChatGPT (version 5.1) to improve the language and readability of the text. In addition, Claude (Opus 5.5) was used to assist in creating the pipeline flowchart (Figure~\ref{fig:pipeline}) and the graphical abstract. After using these tools, the authors reviewed and edited the content as needed and take full responsibility for the content of the publication.





\bibliographystyle{elsarticle-num}
\bibliography{references}  

\end{document}